\documentclass{article} 
\usepackage{iclr2025_conference,times}

\usepackage[utf8]{inputenc} 
\usepackage[T1]{fontenc}    
\usepackage[hidelinks]{hyperref}       
\usepackage{diagbox}
\usepackage{tabularx}
\usepackage{url}            
\usepackage{booktabs}       
\usepackage{amsfonts}       
\usepackage{nicefrac}       
\usepackage{microtype}      
\usepackage{xcolor}       
\usepackage{enumitem}
\usepackage{amssymb}
\usepackage{multirow}
\usepackage{amsmath}
\usepackage{wrapfig}
\usepackage{amssymb}
\usepackage{adjustbox}
\usepackage{arydshln}
\usepackage{dsfont}
\usepackage{subcaption}
\usepackage{graphicx}
\usepackage{amsthm}
\usepackage{array}
\usepackage{etoolbox}
\usepackage{thm-restate}
\usepackage[linesnumbered,ruled,vlined]{algorithm2e}

\usepackage[capitalize,noabbrev]{cleveref}
\usepackage{bm}
\usepackage{amsmath}
\usepackage{rotating}
\usepackage{wrapfig}
\usepackage{mathtools}
\usepackage{caption}

\renewcommand{\citet}{\citep}

\usepackage{color}
\usepackage{enumerate}
\usepackage{graphicx}
\usepackage{amsfonts, amsmath, bm, amssymb}
\usepackage{dsfont}

\usepackage{xspace}
\makeatletter
\DeclareRobustCommand\onedot{\futurelet\@let@token\@onedot}
\def\@onedot{\ifx\@let@token.\else.\null\fi\xspace}

\makeatother

\newcommand{\thetav     }{\boldsymbol \theta     }

\usepackage{amsmath,amsfonts,bm}

\def\eqref#1{equation~\ref{#1}}

\def\1{\bm{1}}

\DeclareMathAlphabet{\mathsfit}{\encodingdefault}{\sfdefault}{m}{sl}
\SetMathAlphabet{\mathsfit}{bold}{\encodingdefault}{\sfdefault}{bx}{n}

\DeclareMathOperator*{\argmin}{arg\,min}

\title{Sharpness-aware Parameter Selection for Machine Unlearning}

\author{Saber Malekmohammadi\thanks{Work done while visiting Emory university} 
\\
David R. Cheriton School of Computer Science\\
University of Waterloo\\
Waterloo, Ontario, Canada \\
\texttt{\{saber.malekmohammadi\}@uwaterloo.ca} \\
\And
Hong kyu Lee \& Li Xiong \\
Department of Computer Science \\
Emory University\\
Atlanta, Georgia, USA \\
\texttt{\{hong.kyu.lee,lxiong\}@emory.edu} 
}

\iclrfinalcopy 
\begin{document}

\maketitle

\begin{abstract}
It often happens that some sensitive personal information, such as credit card numbers or passwords, are mistakenly incorporated in the training of machine learning models and need to be removed afterwards. The removal of such information from a trained model is a complex task that needs to partially reverse the training process. There have been various machine unlearning techniques proposed in the literature to address this problem. Most of the proposed methods revolve around removing individual data samples from a trained model. Another less explored direction is when features/labels of a group of data samples need to be reverted. While the existing methods for these tasks do the unlearning task by updating the whole set of model parameters or only the last layer of the model, we show that there are a subset of model parameters that have the largest contribution in the unlearning target features. More precisely, the model parameters with the largest corresponding diagonal value in the Hessian matrix (computed at the learned model parameter) have the most contribution in the unlearning task. By selecting these parameters and updating them during the unlearning stage, we can have the most progress in unlearning. 
We provide theoretical justifications for the proposed strategy by connecting it to sharpness-aware minimization and robust unlearning. We empirically show the effectiveness of the proposed strategy in improving the efficacy of unlearning with a low computational cost.
\end{abstract}

\section{Introduction}

In the last decade, machine learning has advanced considerably in handling personal data and developing data-driven services. However, the models powering these technologies can unintentionally jeopardize privacy by memorizing sensitive information from training datasets, which could later be exposed to model users~\citep{song2017machine}. For example, \cite{carlini2019secretsharerevaluatingtesting}
revealed that Google’s text prediction tool might autofill credit card numbers appearing in its train data. Moreover, privacy regulations like the European GDPR \citep{GDPR} give people the “right to be forgotten”, enabling them to demand the deletion of their personal data from digital records, which extend to machine learning models trained using the records. Similarly, there are often scenarios where harmful or poisoned data are detected in the train data of a machine learning model and their effect on the trained model needs to be removed after training \citep{warnecke2023}.

Removing specific data records from a machine learning model requires selectively undoing parts of the training process. A naive solution  is to retrain the model entirely without the records to be forgotten, which is both expensive and only feasible if the original data is still accessible. To address this challenge, researchers such as \cite{Cao2015, bourtoule2021} introduced sample unlearning methods, which allow the model to partially reverse its learning and effectively remove certain train data points. This makes it possible to reduce privacy risks and comply with user requests for data removal. In addition, data leaks 
may originate from sensitive features or labels within the dataset, necessitating the removal of their influence from the trained models. 
\citep{warnecke2023}. In such cases, feature or label unlearning allows for targeted elimination of these attributes across multiple samples, rather than sample unlearning.  For example, a sensitive credit card number or password might appear in multiple emails (samples) in a thread of emails. Unlearning this sensitive information requires removing the sensitive information from the group of emails (instead of removing all affected samples from the train set). 

The unlearning scenarios discussed above have been studied in the context of sample/feature/label unlearning \citep{Cao2015, bourtoule2021, guo2023certified, ginart2019makingaiforgetyou, neel2020descenttodeletegradientbasedmethodsmachine, Aldaghri_2021, warnecke2023}. Most existing methods  suffer from high computational complexity or low unlearning performance, due to their approach of updating all model parameters for unlearning (or a naively chosen subset, e.g., the parameters of the last layer of the model), which is expensive and/or inefficient \citep{warnecke2023}. Accordingly, an efficient parameter selection strategy is crucial for performing the unlearning task at various levels of the training data.

In this work, we show that a subset of model parameters have the largest contribution when performing the unlearning task and propose a sharpness-aware (SA) parameter selection strategy for machine unlearning. More specifically, by leveraging second order information about the training loss landscape at the learned model parameters, we find those parameters that are located in a wide area of the loss landscape and update them during unlearning process. By only updating these salient parameters we achieve better unlearn efficacy and better efficiency. We provide justifications for the proposed strategy by connecting it to robust unlearning as well as an approximation of an existing unlearning algorithm designed based on influence functions \citep{warnecke2023, inffunction}.

We summarize our contributions as follows:

\begin{itemize}
    \item We propose a sharpness-aware parameter selection strategy to find the most important model parameters to update during the unlearning procedure. 
    \item We provide theoretical justifications for the proposed strategy by connecting it to sharpness-aware minimization and robust unlearning.

    \item We show the effectiveness of the proposed strategy in improving the efficiency and efficacy of unlearning with a low computational cost. 
\end{itemize}
\section{Related work}

As previously mentioned, training of machine learning models often demands a high volume of data, which might contain privacy-sensitive information, and lead to privacy leakage \citep{cristofaro, papernotsok}. Accordingly, detecting and correcting privacy issues of machine learning models has been the topic of a recent research direction \citep{carlini2019secretsharerevaluatingtesting, Zanella_B_guelin_2020, carlini2021extractingtrainingdatalarge,
salem2018mlleaksmodeldataindependent,
Leino2019StolenML,
shokri2017membershipinferenceattacksmachine}. Methods for removing sensitive data from learning models are a recent branch of security research. As one of the first, \cite{Cao2015} showed that a large number of learning models can be represented in a closed summation form that allows elegantly removing individual data points in retrospection. However, for adaptive learning strategies, such as stochastic gradient descent, this approach provides only little advantage over retraining and thus is not well suited for correcting problems in deep neural networks.

\cite{bourtoule2021} addressed this problem and proposed exact unlearning, a strategy that completely unlearns data instances from general classification models. Similarly, \cite{ginart2019makingaiforgetyou} developed a technique for unlearning points from clustering algorithms. The key idea of both approaches is to split the data into independent partitions— so called shards—and aggregate the final model from the sub-models trained over these shards. Due to this partitioning of the model, unlearning data points can be efficiently realized by retraining the affected sub-models only, while the remaining sub-models remain unchanged. \cite{Aldaghri_2021} showed that this approach can be further sped up for least squares regression by choosing the shards cleverly. We refer to this family of unlearning methods as sharding. Unfortunately, sharding has one critical drawback: Its efficiency quickly deteriorates when multiple data points need to be corrected. The probability that all shards need to be retrained increases with the number of affected data points \citep{warnecke2023}. To address this inefficiency, various approaches have been used to increase the efficiency of exact machine unlearning frameworks. \cite{yan2022arcane} proposed ARCANE, a more efficient unlearning algorithm based on~the work by \cite{bourtoule2021}: Instead of random partitioning, the authors have partitioned the data with respect to their classes. Experimental results demonstrated increased efficiency and model utility.

Another strain of machine unlearning is approximate unlearning, which usually achieves unlearning by conducting several gradient updates. The most straightforward approach is gradient ascent over target data~\citep{golatkar2020eternal}. Several works discuss efficiency of the approximate unlearning. \cite{jia2022model} showed that conventional model sparsification strategies, such as weight pruning, improve machine unlearning. Specifically, the authors demonstrated that a model unlearned after pruning exhibited smaller performance gap with the retrained model. Along with this finding, the authors proposed machine unlearning frameworks aided by sparsification. Recently, \cite{li2025funu} proposed an efficient unlearning by reducing the number of data samples to unlearn. The authors identified that certain samples do not fundamentally change the model behavior after unlearning. Thus, by filtering out such unnecessary data, the authors are able to efficiently achieve unlearning by only removing crucial data samples.

In the context of feature/label unlearning, considering the limitations of sample unlearning methods, \cite{warnecke2023} introduced a method for unlearning features and labels. Their proposed method exploits the concept of influence functions to assess the impact of a change in each train data point on the trained model. Using the assessed impact of a change in the train data, they propose a direct one-step update to model parameters for unlearning. However, their proposed method updates all model parameters (or a naively chosen subset), which makes it computationally expensive (suboptimal), especially when the size of the used model increases. 

For large language models, increasing efficiency of unlearning frameworks is even more crucial due to their large number of parameters. Similar to fine-tuning, LoRA~\citep{hu2022lora} is widely used for unlearning large language models. \cite{gao2024practical} proposed a practical unlearning algorithm for large language models. It freezes the original weights and fine-tunes LoRA with target prompts to deter the model from providing the original response. Then, it utilizes a classification model to determine the similarity of given prompts with the unlearning contents. Depending on the similarity, adapters are activated to emit the safe answers. Similarly, \cite{cha2024towards} utilized gradient ascent with LoRA. Instead of directly utilizing gradient ascent on LoRA, the authors further proposed additional hinge loss to boost unlearning stability. \cite{hu2024exact} proposed exact unlearning on LLMs. Instead of training a separate model on each data partition, the authors also partition adapters and train each adapter partition on each corresponding data partition. Recently, \cite{jia2025wagle} proposed a parameter selection strategy for large language model unlearning. The authors score the model weights  by their gradient sensitivity. 
Specifically, they introduce small changes to the gradients, then select the weights that demonstrate significant response to the change, which are considered critical for unlearning. However, this requires solving complex optimization problems, which could lower the overall efficiency.

Hessians, or the second order gradients are widely utilized in machine unlearning. Primarily, certified unlearning~\citep{guo2019certified} requires the inverse of Hessians to analyze the gradient norm and bounding unlearning budget, 
which is expensive to compute regardless of the complexity of the model~\citep{zhang2024towards}. Some works utilized Hessian approximations to increase efficiency~\citep{Mehta_2022_CVPR}. Unlike our work, the primarily focus of Hessians in certified unlearning is to complete the mathematical analysis, rather than finding the salient parameters.
Hessians are rarely utilized for salient parameter selection. \cite{liu2023unlearning} used the fisher information matrix, or expected hessian matrix to selectively choose parameters to update. The authors utilize influence function and derive upper bound of Kullback-Leibler divergence between original parameters and estimated retrained parameters. The upper bound is determined by diagonals of the fisher information matrix. The authors choose parameters that can lower the upper bound. However, although the paper empirically show their approach is effective for deep models, the upper bound analysis is only valid to linear models. Some works conduct gradient ascent assisted by inverse Hessian matrix, to accelerate optimization of parameters containing knowledge of unlearning samples from large language models~\citep{jia2024soul, gu2024second}. Most of them utilized the hessian matrix in the broad context of second order update. While few works choose parameters via fisher information matrix, and some attempted parameter selection via first order gradient~\citep{fan2023salun}. Recently, ~\cite{foster2024fast} proposed selective synaptic dampening (SSD), a salient parameter selection using fisher information matrix. The authors obtain two Fisher information matrices: one from the retain data and one from the forget data. They then select the parameters whose corresponding elements in the Fisher information matrix from the forgotten data are higher than those in the matrix from the retained data.~\cite{NEURIPS2024_2e622ac7} proposed unlearning with steepest descent and also used the same selection strategy. While this is closely related to our work, it is not directly applicable to our problem settings. SSD expects the sample unlearning scenario (unlearning both features and labels) while our problem setting is the feature unlearning (unlearn a subset of features or labels).


\section{Notations and background}

Let $x\in \mathcal{X}\subseteq\mathbb{R}^d$ and $y \in \mathcal{Y}=\left\{1, \ldots, C \right\}$ denote an input data point and its target label. Let us consider a train dataset $\mathcal{D} = \mathcal{D}_r \cup \mathcal{D}_f = \{z_i\}_{i=1}^N$ with $N$ samples $z_i = (x_i, y_i)$. The subset $\mathcal{D}_r$ includes the train samples to be retained and $\mathcal{D}_f$ is the set of train samples with a subset of their features/labels needed to be reverted and hence have to be unlearned (forget set). Let $h: \mathcal{X}\times \mathbf{\thetav} \to\mathbb{R}^C$ be the predictor function, which is parameterized by $\mathbf{\thetav}\in \mathbb{R}^p$. Also, let $\ell:\mathbb{R}^C\times\mathcal{Y}\to \mathbb{R}_+$ be the used loss function (cross-entropy loss). We also define empirical train loss as $f_S(\mathbf{\thetav})=\frac{1}{N}\sum_{(x,y)\in \mathcal{D}}[\ell(h(x,\mathbf{\thetav}), y)]$, with minimum value $f^*$. 

\subsection{Sharpness-Aware Minimization (SAM)}\label{sec:sam}
Our proposed parameters selection strategy is inspired by SAM. We provide a brief introduction about it in this section. SAM is an optimization technique that  minimizes both the loss value and its sharpness, leading to those solution points that not only have a small loss value but also are located in a flat region of the optimization landscape, enhancing generalization performance and scalability of the trained model \citep{foret2021sharpnessawareminimizationefficientlyimproving, andriushchenko2022understandingsharpnessawareminimization, liu2022efficientscalablesharpnessawareminimization, du2023sharpnessawaretrainingfree, zhang2023missingirmtrainingevaluation}. 


Minimizing the empirical loss $f_S$ using an optimization procedure, e.g. SGD or Adam, is a common approach for learning models on a train set $\mathcal{D}$. In practice, there is usually a distribution shift between the train set and the test set and optimizing an over-parameterized model, e.g. deep neural networks, on a given train set results in suboptimal performance at test time, i.e. there is a generalization gap for the model parameters found by minimizing $f_{S}$. \cite{foret2021sharpnessawareminimizationefficientlyimproving} showed that it is not only the value of $f_S(\thetav)$ that determines the generalization gap of a model parameter $\thetav$, but also there is a relation between the sharpness of the loss landscape at the found parameter $\thetav$ and its generalization gap. Accordingly, SAM was proposed to find those model parameters whose entire neighborhood has a low empirical loss $f_S$, rather than model parameters with only a small empirical loss value. In other words, SAM looks for those model parameters with both low loss value and low curvature. Instead of simply minimizing $f_S(\thetav)$, SAM aims at minimizing the following sharpness-aware loss function with a neighborhood of size $\rho$:

\begin{align}\label{eq:SAM}
    \min_{\thetav} f_S^{SAM}(\thetav) + \lambda ||\thetav||_2^2, ~~~~~~\textit{where} ~~~~~ f_S^{SAM}(\thetav) := \max_{||\epsilon||_2\leq \rho} f_S(\thetav + \epsilon).
\end{align}

The above objective function of SAM prevents the model from converging to a sharp minimum in the loss landscape, leading to more generalizable solutions \citep{foret2021sharpnessawareminimizationefficientlyimproving}.

\section{Sharpness-Aware Parameter Selection for Unlearning}
\textbf{Problem Definition}. Let us recall the supervised learning setting introduced earlier with the data set $\mathcal{D}=\{z_i\}_{i=1}^N$, where the sample point $z_i = (x_i, y_i)$ is a pair of features $x_i \in \mathbb{R}^d$ and the corresponding label $y
_i$. Let us define the following loss function:

\begin{align}\label{eq:L_learn}
    L_{\textit{learn}}(\thetav; \mathcal{D}):=
    \sum_{(x_i,y_i)\in \mathcal{D}}\ell(h(x_i,\mathbf{\thetav}), y_i) =  \underbrace{\sum_{(x_i,y_i)\in \mathcal{D}_f}\ell(h(x_i,\mathbf{\thetav}), y_i)}_{L_{\mathcal{D}_f}} + \underbrace{\sum_{(x_i,y_i)\in \mathcal{D}_r}\ell(h(x_i,\mathbf{\thetav}), y_i)}_{L_{\mathcal{D}_r}}
\end{align}
Therefore, the optimum model parameter $\thetav^* \in \mathbb{R}^p$ can be found from the following optimization problem:

\begin{align}\label{eq:learn}
    \thetav^* = \argmin_{\thetav} L_{\textit{learn}}(\thetav; \mathcal{D}).
\end{align}

Let us consider a scenario where some data points in $\mathcal{D}$, denoted as set $\mathcal{D}_f$, have been changed after learning $\thetav^*$ above. For instance, a point $z_i=(x_i, y_i) \in \mathcal{D}_f$ has been changed to $\Tilde{z}_i = (\Tilde{x}_i, \Tilde{y}_i)$ by changing either its feature vector and/or its label. Let $\tilde{\mathcal{D}}_f$ denote the set of changed sample points and $\mathcal{D}_r$ denote the remaining set of points in $\mathcal{D}$ that have not been changed. Accordingly, we define the following loss function:

\begin{align}\label{eq:L_unlearn}
    L_{\textit{unlearn}}(\thetav; \mathcal{D}_r, \tilde{\mathcal{D}}_f) &= \sum_{(x_i,y_i)\in \mathcal{D}_r}\ell(h(x_i,\mathbf{\thetav}), y_i) + \sum_{(\tilde{x}_i,\tilde{y}_i)\in \tilde{\mathcal{D}}_f}\ell(h(\tilde{x}_i,\mathbf{\thetav}), \tilde{y}_i) \nonumber \\
    &=L_{\textit{learn}}(\thetav; \mathcal{D}_r, \mathcal{D}_f) - \sum_{(x_i,y_i)\in \mathcal{D}_f}\ell(h(x_i,\mathbf{\thetav}), y_i) + \sum_{(\tilde{x}_i,\tilde{y}_i)\in \tilde{\mathcal{D}}_f}\ell(h(\tilde{x}_i,\mathbf{\thetav}), \tilde{y}_i).
\end{align}

Note that setting $\tilde{\mathcal{D}}_f=\emptyset$ in \Cref{eq:L_unlearn} yields to removing the effect of data samples in $\mathcal{D}_f$. Hence, the above loss function can be used for a general unlearning scenario, including sample/feature/label unlearning. Having the forget set $\mathcal{D}_f$ and new set $\tilde{\mathcal{D}}_f$, the question is how can we update the parameter $\thetav^*$, which was learned on $\mathcal{D}_r$ and $\mathcal{D}_f$, into a new solution $\tilde{\thetav}^*$ as if it was learned on $\mathcal{D}_r$ and $\tilde{\mathcal{D}}_f$? Ideally, we have to find the solution to the following optimization problem:

\begin{align}\label{eq:unlearn_scratch}
    \tilde{\thetav}^* = \argmin_{\thetav} L_{\textit{unlearn}}(\thetav; \mathcal{D}_r, \tilde{\mathcal{D}}_f).
\end{align}

If we can afford the computational cost, the unlearning problem \ref{eq:unlearn_scratch} can be solved by retraining the model from scratch on $\mathcal{D}_r \cup \tilde{\mathcal{D}}_f$. However, retraining from scratch is expensive and instead, we look for an unlearning algorithm to update the parameter $\thetav^*$ to $\tilde{\thetav}^*$ (or an estimate of it) to reduce the cost of unlearning $\mathcal{D}_f$. We can formulate the unlearning task as an alternative optimization problem. Intuitively, the unlearned model should remove the effect of the samples in $\mathcal{D}_f$, while preserving the performance of the model on $\mathcal{D}_r$ (and ensuring a satisfying performance on $\tilde{\mathcal{D}}_f$). Accordingly, the unlearning task is usually modeled as solving the following alternative optimization problem \emph{starting from the pretrained parameters $\thetav^*$}:

\begin{align}\label{eq:unlearn_model}
    \thetav^u \approx \argmin_{\thetav} L_{\textit{forget}}(\thetav; \mathcal{D}_f, \tilde{\mathcal{D}}_f) + \lambda L_{\mathcal{D}^\prime_r}(\thetav; \mathcal{D}^\prime_r).
\end{align}

The regularization weight $\lambda$ balances the forget and retain tasks and $\mathcal{D}^\prime_r$ is usually a subset of the retain set $\mathcal{D}_r$. The regularization loss term is for maintaining a good performance on the retain set $\mathcal{D}_r$.The solution $\thetav^u$ is an approximation of the true unlearning solution $\tilde{\thetav}^*$ in problem \ref{eq:unlearn_scratch}. The forget loss $L_{forget}$ is designed such that it forgets the set $\mathcal{D}_f$ and learns the new set $\tilde{\mathcal{D}}_f$. Accordingly, a common approach for designing $L_{forget}$ is to treat the samples in $\mathcal{D}_f$ ($\tilde{\mathcal{D}}_f$) as negative (positive) samples \citep{zhang2024negativepreferenceoptimizationcatastrophic, rafailov2024directpreferenceoptimizationlanguage}. For instance:

\begin{align}
    L_{forget}(\thetav; \mathcal{D}_f, \tilde{\mathcal{D}}_f) := \sum_{(\tilde{x}_i,\tilde{y}_i)\in \tilde{\mathcal{D}}_f}\ell(h(\tilde{x}_i,\mathbf{\thetav}), \tilde{y}_i) - \sum_{(x_i,y_i)\in \mathcal{D}_f}\ell(h(x_i,\mathbf{\thetav}), y_i).
\end{align}

As observed, the objective function in problem \ref{eq:learn} (of the learning task) and objective function in problem \ref{eq:unlearn_model} (of the unlearning task) have a common part (both $L_{\mathcal{D}_r}$ and $L_{\mathcal{D}^\prime_r}$ target a satisfying performance on $\mathcal{D}_r$). However, the terms $L_{\mathcal{D}_f}$ in problem \ref{eq:learn} and $L_{\textit{forget}}$ in problem \ref{eq:unlearn_model} are in conflict: the former ensures a good performance on $\mathcal{D}_f$, while the latter tries to forget the set $\mathcal{D}_f$ (and if there is a set $\tilde{\mathcal{D}}_f$, the latter encourages a good performance on it).

\subsection{sharpness-aware parameter selection for unlearning}

An important question is that if, due to a limited computational budget,  we are limited to update only a subset of the model parameters in $\thetav^*$ to unlearn $\mathcal{D}_f$, updating which subset of parameters will lead to a better unlearning performance? In this section, we propose our parameter selection strategy and also provide a justification for it. 

When looking for a method to update $\thetav^*$ to an approximation of $\tilde{\thetav^*}$, there are a critical subset of parameters in $\thetav^*$ that are the most important ones to update. Finding and updating these most important parameters enhances both unlearning efficacy and efficiency at a lower computational cost. Intuitively, in our selection strategy, we look for those parameters in $\thetav^*$ that $L_{\textit{learn}}$ loss landscape is widely flat along their direction. In fact, these parameters are the hardest ones to update during the unlearning procedure because they are located in a wide minimum of the loss landscape and changing them slightly would not incur a big change in $L_{\textit{learn}}$. Hence, we propose a sharpness-aware parameter selection strategy. The sharpness of the $L_{\textit{learn}}$ landscape in  a neighborhood of size $\rho$ around the found solution $\thetav^*$ can be measured as:

\begin{align}
    \max_{||\epsilon||\leq \rho} L_{\textit{learn}}(\thetav^* + \epsilon; \mathcal{D}_r, \mathcal{D}_f) - L_{\textit{learn}}(\thetav^*; \mathcal{D}_r, \mathcal{D}_f).
\end{align}

The above measure being not tractable, many sharpness measures in the literature rely on second-order derivative characteristics of $L_{\textit{learn}}$ such as the trace or the operator norm of its Hessian matrix $H_{\textit{learn}}$ at $\thetav^*$ \citep{chaudhari2017entropysgdbiasinggradientdescent, keskar2017largebatchtrainingdeeplearning}, which can be found with complexity $\mathcal{O}(p)$. As our sharpness-aware parameter selection strategy for updating the found solution $\thetav^*$ through the unlearning task, we propose to update those parameters (directions) in $\thetav^*$ with the lowest corresponding loss sharpness or equivalently, those with the smallest corresponding diagonal value in $H_{\textit{learn}}(\thetav^*)$. In other words, we choose to update those parameters of $\thetav^*$ with the lowest sharpness of the $L_{\textit{learn}}$ landscape in their direction. In the following, we provide two theoretical justifications for this strategy.

\subsubsection{sharpness-aware parameter selection strategy through the lens of robust unlearning} 
Robust unlearning seeks for robustness against relearning attacks \citep{hu2024joggingmemoryunlearnedllms}. The attacks aim to recover the knowledge that was unlearned through the unlearning process by fine-tuning the unlearned model $\thetav^u$ using a very small subset $\mathcal{D}'_f$ of the forget set $\mathcal{D}_f$. The relearning attack can be formulated as follows:

\begin{align}
    \min_{\epsilon} L_{relearn}(\thetav^u + \epsilon; \mathcal{D}'_f),
\end{align}

where $L_{relearn}$ behaves in contrast to $L_{forget}$, e.g. it is the negative of $L_{forget}$ on $\mathcal{D}'_f$ or is the standard fine-tuning loss on $\mathcal{D}'_f$ \citep{fan2025llmunlearningresilientrelearning}. In this case, the above relearning attack can be reformulated as:

\begin{align}\label{eq:relearnattack}
    \max_{\epsilon} L_{forget}(\thetav^u + \epsilon; \mathcal{D}'_f).
\end{align}

According to the attack described above, the robust unlearning to learning attacks can be modeled as an adversary-defense game expressed as a min-max optimization such that the robust unlearning process counteracts the adversarial relearning process \citep{cha2024towards}:

\begin{align}\label{eq:robustul}
    \min_{\thetav} \underbrace{\max_{||\epsilon||_p \leq \rho} L_{forget}(\thetav; \mathcal{D}_f, \tilde{\mathcal{D}}_f)}_{L_{forget}^{SAM}(\thetav; \mathcal{D}_f, \tilde{\mathcal{D}}_f)} + \lambda L_{\mathcal{D}_r}(\thetav; \mathcal{D}_r). 
\end{align}

The constraint $||\epsilon||_p \leq \rho$ (with $p=2$, by default) limits the ability of the adversary for manipulating the unlearned model $\thetav^u$. Interestingly, according to equation \ref{eq:SAM}, problem \ref{eq:robustul} is the same as sharpness-aware minimization (SAM) of $L_{forget}$. Intuitively, SAM of $L_{forget}$ yields an unlearned model parameter $\thetav^u$, which is located in a wide dip in the landscape of $L_{forget}$ and is robust to relearning changes by an adversary.

Recall from problem \ref{eq:learn} that $L_{learn}(\thetav; \mathcal{D}_r, \mathcal{D}_f)$ is in contradiction with $L_{forget}(\thetav; \mathcal{D}_f, \tilde{\mathcal{D}}_f)$ in problem \ref{eq:robustul}. In other words, those model parameters in $\thetav^*$ that are in a wide minimum in the landscape of $L_{learn}(\thetav; \mathcal{D}_r, \mathcal{D}_f)$ cannot be in a wide minimum of $L_{forget}(\thetav; \mathcal{D}_f, \tilde{\mathcal{D}}_f)$ simultaneously. Therefore, in order to achieve a robust unlearning, those parameters in $\thetav^*$ that are in a wide minimum in the landscape of $L_{learn}(\thetav; \mathcal{D}_r, \mathcal{D}_f)$ are in priority for getting updated through the unlearning process. 
And these parameters are chosen by our sharpness-aware parameter selection strategy.
This verifies our intuition that by focusing on these parameters that are easier to relearn or harder to unlearn, we will achieve more effective and robust unlearning.


\subsubsection{sharpness-aware parameter selection strategy through the lens of approximate parameter updating}
Another justification of our parameter selection strategy can be provided by the unlearning algorithm proposed by \cite{warnecke2023}. The authors proposed to apply a closed-form update to the parameters of the learned $\thetav^*$ for unlearning. More specifically, they proposed a \emph{one-step} second-order update to $\thetav^*$ based on the forget loss $L_{forget}(\thetav;\mathcal{D}_f, \tilde{\mathcal{D}_f})$, as follows:

\begin{align}\label{eq:secondorder}
    \thetav^u &\approx \thetav^* - H_{learn}^{-1}(\thetav^*) \big(\nabla_{\thetav} L_{forget}(\thetav^*;\mathcal{D}_f, \tilde{\mathcal{D}_f})\big),
\end{align}

\emph{where $H_{learn}(\thetav^*)$ is the Hessian of $L_{learn}(\thetav;\mathcal{D}_r, \mathcal{D}_f)$ at $\thetav^*$}. Therefore the above second-order update is not a Newton update, as the Hessian and the gradient vector are computed on two different losses $L_{learn}$ and $L_{forget}$, respectively. From the above update rule, we can clearly observe that those parameters in $\thetav^*$ which are located in  a wide dip of $L_{learn}$ will experience a larger learning rate through the update process. Therefore, those are the parameters which will be updated the most. Therefore, our proposed parameter selection provides a valid approximation to the above closed-form second-order update for unlearning, by eliminating the need to compute the inverse of the complete Hessian matrix. In fact, our strategy needs only the diagonal elements of the Hessian matrix for performing its parameter selection, which can be computed with computational complexity of $\mathcal{O}(p)$.

\section{Results}
In this section, we report experimental results to show the effectiveness of the proposed parameter selection method. 

\subsection{Experimental Setup}
We consider a classification task and use a convolutional neural network (CNN) to classify samples in MNIST dataset and a ResNet model to classify CIFAR10 samples. \Cref{table:split_uniform} summarizes the models and the datasets. Let us assume we have learned a model parameter $\thetav^*$ on a poisoned dataset by minimizing $L_{learn}$ (problem \ref{eq:learn}). We consider unlearning the poisoned labels by using the proposed parameter selection strategy. For the poisoning attack, we consider a scenario where the labels of 2500 samples (out of the total 50000 train samples) are poisoned and need to be unlearned to improve utility. We report the results on 10000 held-out samples. We consider two scenarios, where the 2500 poisoned samples are all from the same class (targetted attack) or from different classes (untargetted attack). 

In our method, we select parameters based on their corresponding diagonal elements in the Hessian matrix of $L_{learn}$ at $\thetav^*$. Let us assume we want to update only one of the model layers for performing unlearning. Then, we update the layer of the model for which the average Hessian diagonal value is the largest among all layers. 

We compare our method with two baselines.  The first one uses the method proposed by \cite{warnecke2023}, which is an unlearning approach that translates changes in the training data into direct updates to the model parameters using a one-step first-order closed-form update. The baseline uses a naive parameter selection strategy of updating only the parameters in the last layer which is in contrast to ours. Fine-tuning all parameters on only $\mathcal{D}_r$ for some epochs is the second baseline that we compare with.

\begin{table*}[t]
\centering
\caption{Details of the experiments and the used datasets. ResNet-18 are the residual neural networks defined in \cite{resnet}. CNN: Convolutional Neural Network. }
\label{tab:datasets}
\small
\setlength\tabcolsep{2pt}
\begin{tabular}{ccccccc}
\toprule
\bf{Datasets} & \bf{Train set size} & \bf{Test set size} & \bf{Model} & \bf{\# of parameters} 
\\ 
\midrule
MNIST & 50000 & 10000 & CNN & 28,938\\

CIFAR10 & 50000 & 10000 & ResNet-18 (\cite{resnet}) & 11,181,642
\\

\bottomrule
\end{tabular}
\label{table:split_uniform}
\end{table*}

\begin{figure}[t]
    \centering
    \includegraphics[width=0.36\columnwidth,height=3.5cm]{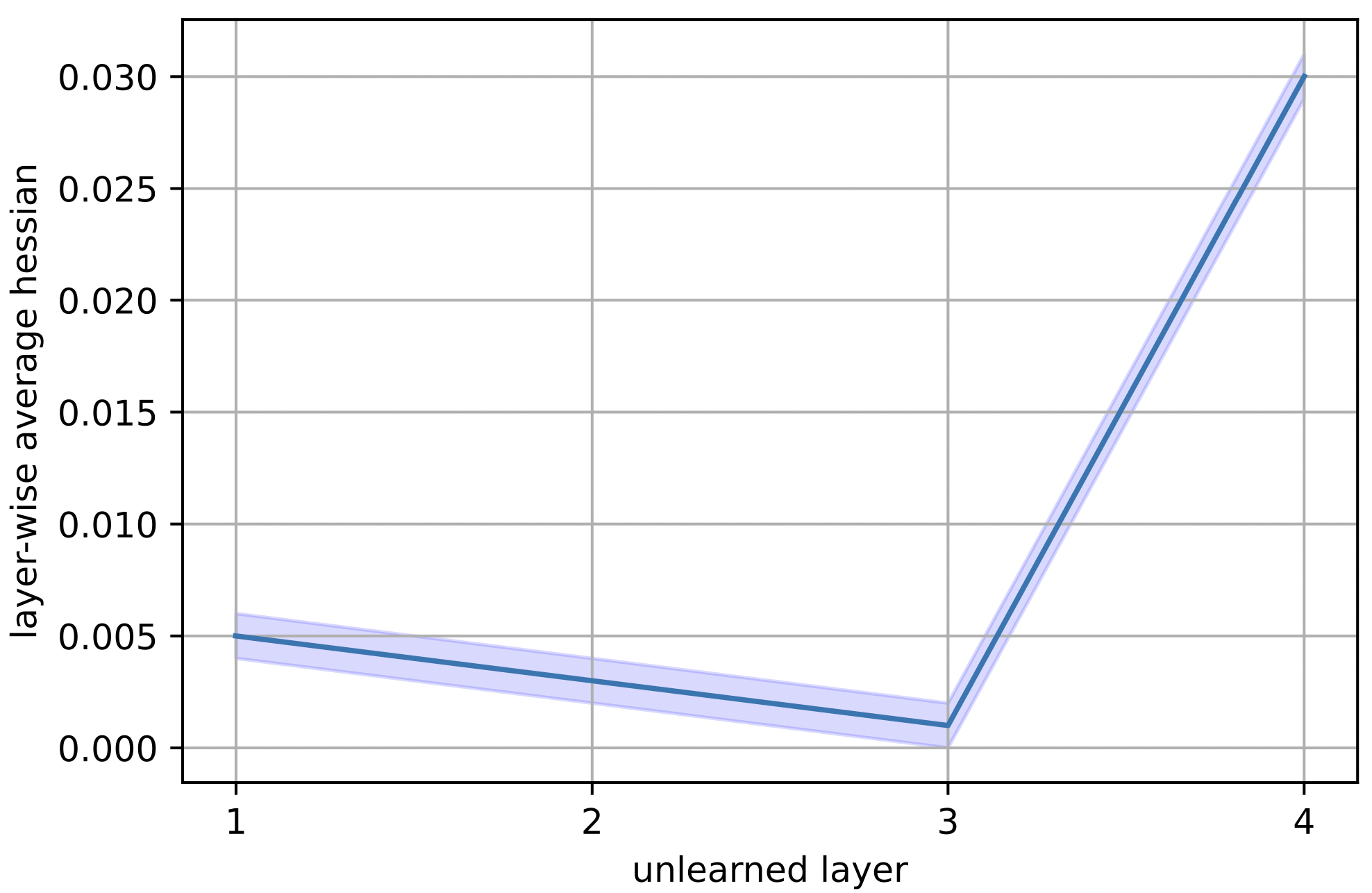}
    ~~~~\includegraphics[width=0.36\columnwidth,height=3.5cm]{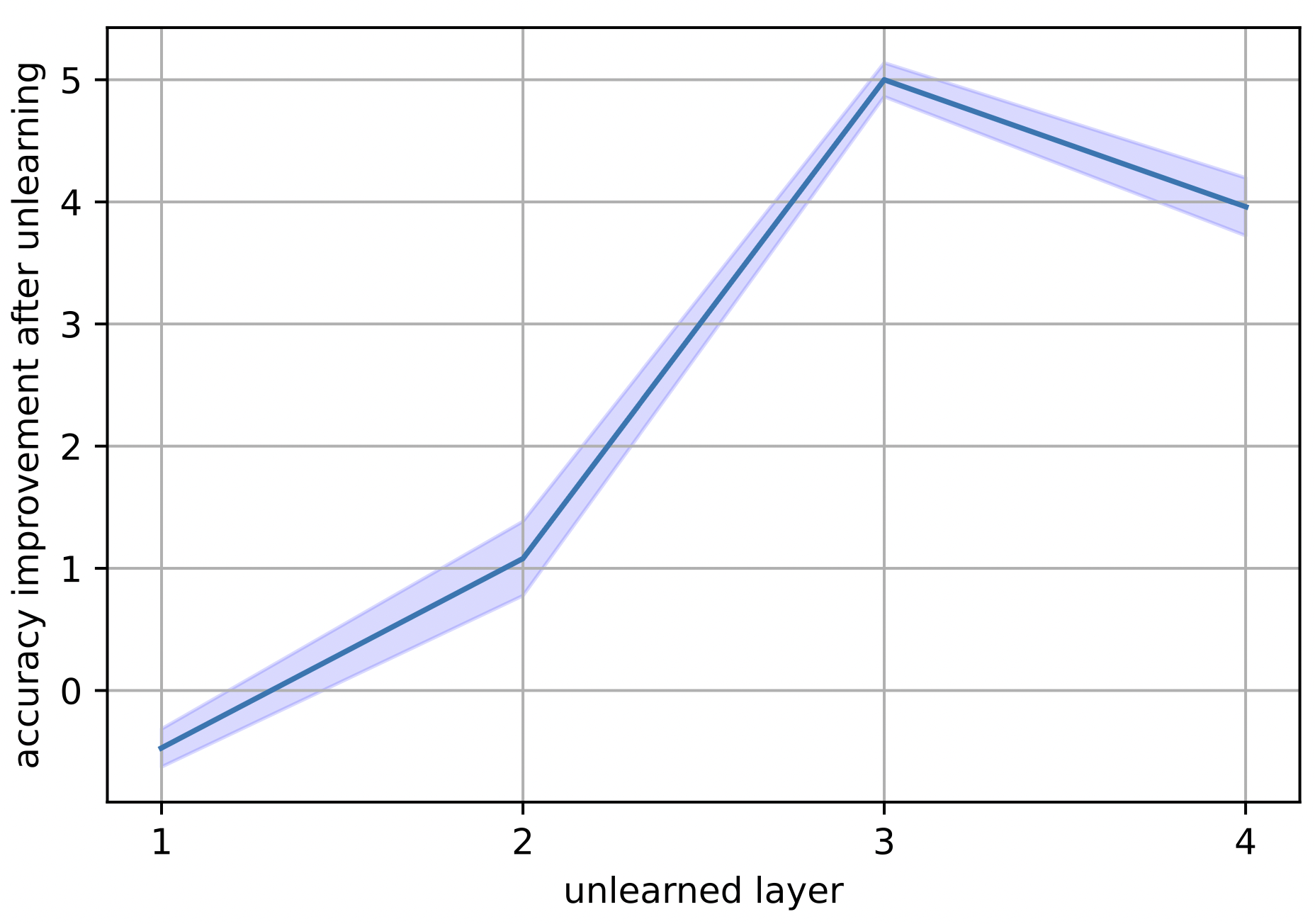}
    
    \caption{\textbf{Left:} the average sharpness (diagonal Hessian) of different layers of the model learned on poisoned data. \textbf{Right:} the average test accuracy improvement after performing unlearning on the selected layer (when the other layers are frozen). As observed, the layer with smallest average sharpness results in the largest improvement compared to other layers, including the last layer.}
    \label{fig:paramselection}
\end{figure}

\subsection{Results}
\Cref{fig:paramselection} left, on MNIST dataset, compares the layer-wise average diagonal Hessian value on $L_{learn}$ for different layers of the model at $\thetav^*$. The figure on the right compares the amount of improvement in the utility over the held-out data when each of the 4 layers is used for unlearning untargetted poisoned labels with the first-order closed-form update of \citep{warnecke2023} mentioned before. As observed, updating layer 3, which has the lowest layer-wise average diagonal Hessian value results in the largest utility improvement over the held-out data. This shows the effectiveness of our strategy compared to the first baseline, which naively chooses the last layer of the model for performing the unlearning task with the same first-order parameter update. This observation proves that the parameters with the smallest sharpness of the loss landscape along their direction are the most important ones to update for unlearning.

\begin{table*}[hbt!]
\centering
\caption{Results on CIFAR10 showing accuracy on untargetted classes (all classes except class 9) in held-out data after unlearning targetted poisoned samples.}
\label{tab:targetted_untargetted}
\small
\setlength\tabcolsep{2pt}
\begin{tabular}{cccc}
\toprule
\bf{Poisoned samples} & \bf{Fine-tuning} & \bf{1st order (last layer)} & \bf{SA 1st order}  
\\ 
\midrule
2500 & 82.12\scriptsize $\pm$ 0.91 & 85.34\scriptsize $\pm$ 0.74 & \textbf{86.71\scriptsize $\pm$ 1.2}
\\
\bottomrule
\end{tabular}
\end{table*}

\begin{table*}[hbt!]
\centering
\caption{Results on CIFAR10 showing accuracy on the targetted calss (class 9) in held-out data after unlearning targetted poisoned samples.}
\label{tab:targetted_targetted}
\small
\setlength\tabcolsep{2pt}
\begin{tabular}{cccc}
\toprule
\bf{Poisoned samples} & \bf{Fine-tuning} & \bf{1st order (last layer)} & \bf{SA 1st order}  
\\ 
\midrule
2500 & 72.02\scriptsize $\pm$ 1.1 & 74.31\scriptsize $\pm$ 0.98 & \textbf{75.61\scriptsize $\pm$ 0.84}
\\
\bottomrule
\end{tabular}
\end{table*}

The results in \Cref{tab:targetted_untargetted} on CIFAR10 compare the proposed approach for unlearning targetted (class 9) poisoning attacks with the two baselines.  
All the methods are compared in terms of the test accuracy on the ``untargtted" classes (including all classes except class 9) held-out data to show how effective the methods are in terms of retaining the performance on the untargetted samples (fidelity of unlearning).

As observed the proposed parameter selection strategy is more effective than when unlearning is performed on the last layer of the model or when all model parameters are updated using fine-tuning. 

It is also noteworthy that fine-tuning all model parameters for several epochs on $\mathcal{D}_r$ enforces even larger computational cost than the proposed method. In fact, the proposed method performs only one-step first-order update (going over the whole samples in $\mathcal{D}_f$ and $\tilde{\mathcal{D}}_f$ at once) to only the selected set of parameters. 

Similarly, the results reported in \Cref{tab:targetted_targetted} completes the previous table and compares the same algorithms based on the test accuracy on the ``targetted" class (class 9) after performing unlearning. This results evaluate the unlearning methods based on how effective they unlearn the targetted poisoned samples (efficacy of unlearning).

\begin{table*}[t!]
\centering
\caption{Results on CIFAR10 showing accuracy on held-out (validation) data after unlearning untargetted poisoned samples.}
\small
\setlength\tabcolsep{2pt}
\begin{tabular}{cccccc}
\toprule
\bf{Poisoned samples} & \bf{Fine-tuning} & \bf{1st order (last layer)} & \bf{SA 1st order}
\\ 
\midrule
2500 & 84.02\scriptsize $\pm$ 0.82 & 85.21\scriptsize $\pm$ 1.02 & \textbf{86.71\scriptsize $\pm$ 0.65}\\

5000 &  83.04\scriptsize $\pm$ 1.2 & 84.34\scriptsize $\pm$ 0.92 & \textbf{85.60\scriptsize $\pm$ 1.1}
\\
\bottomrule
\end{tabular}
\label{tab:untargetted_N_poison}
\end{table*}

As observed, the proposed approach is the most effective one in forgetting the poisoned labels, thanks to its parameter selection strategy and updating the most important parameters. In contrast, the other two methods achieve a lower accuracy on the targetted class (lower efficacy) with the same or a larger computational cost.  In summary, our sharpness aware (SA) parameter selection strategy enhances unlearning in the two critical dimensions: improving unlearning efficacy, preserving model utility on non-unlearned tasks.  

Finally, in \Cref{tab:untargetted_N_poison}, on CIFAR10 dataset, we have compared the same algorithms above based on an ``untargetted" poisoning attack for different number of poisoned samples. We have considered the cases where the lables of 2500 and 5000 samples (out of 50000 total train samples) are randomly flipped. 
We observe that, in both the scenarios, the proposed approach performs better than fine-tuning all parameters and naively updating only the last layer of the model.


\section{Conclusion}

Most existing sample/feature/label unlearning methods perform unlearning by updating the whole set of model parameters or only a naively chosen subset of them. This leads to a high computational cost and/or low unlearning performance. As a remedy, we proposed a parameter selection strategy for performing the unlearning task. The selection approach is sharpness-aware in the sense that it chooses the parameters which have the lowest sharpness of the learning loss landscape along their direction. We showed how this selection strategy approximates an expensive optimal second-order update and how it facilitates achieving robustness to relearning attacks. In our future ongoing research, we will enhance the theoretical and experimental results developed in this work on larger models and large language models (LLMs).






\clearpage
\newpage

\bibliography{iclr2025_conference}
\bibliographystyle{iclr2025_conference}


\end{document}